\documentclass[11pt,a4paper]{article}
\usepackage[hyperref]{emnlp2018}
\usepackage{times}
\usepackage{latexsym}

\usepackage{url}
\usepackage{graphicx}
\usepackage{multirow}
\usepackage{subcaption}
\usepackage{caption}
\usepackage{float}
\usepackage{tikz}
\usepackage{amsmath}
\usepackage{mathrsfs}
\usepackage{multirow}
\usepackage{color}
\usepackage{ulem}
\usepackage{natbib}
\usepackage{CJK}

\aclfinalcopy % Uncomment this line for the final submission
\newcommand*{\affaddr}[1]{#1} % No op here. Customize it for different styles.

\newcommand*{\email}[1]{\texttt{#1}}

\title{Review-Driven Multi-Label Music Style Classification \\by Exploiting Style Correlations}

\author{
Guangxiang Zhao\thanks{\ \ Equal Contribution}, \ 
Jingjing Xu\footnotemark[1], \ 
Qi Zeng, \
Xuancheng Ren
\\ 
\affaddr{MOE Key Lab of Computational Linguistics, School of EECS, Peking University}  \\
\email{\{zhaoguangxiang,jingjingxu,pkuzengqi,renxc\}@pku.edu.cn }\\
}

\date{}

\begin{document}
\maketitle

\begin{abstract}
  This paper explores a new natural language processing task, review-driven multi-label music style classification. This task requires the system to identify multiple styles of music based on its reviews on websites. The biggest challenge lies in the complicated relations of music styles. It has brought failure to many multi-label classification methods. To tackle this problem, we propose a novel deep learning approach to automatically learn and exploit style correlations. The proposed method consists of two parts: a label-graph based neural network, and a soft training mechanism with correlation-based continuous label representation. Experimental results show that our approach achieves large improvements over the baselines on the proposed dataset. Especially, the micro F1 is improved from 53.9 to 64.5, and the one-error is reduced from 30.5 to 22.6. Furthermore, the visualized analysis shows that our approach performs well in capturing style correlations.% to a certain extent.
\end{abstract}
\begin{CJK}{UTF8}{gbsn}
\section{Introduction}

%介绍任务 什么是音乐类别。解释说明，为什么要有这个任务。
As music style (e.g., Jazz, Pop, and Rock) is one of the most frequently used labels for music, music style classification is an important task for applications of music recommendation, music information retrieval, etc. There are several criteria related to the instrumentation and rhythmic structure of music that characterize a particular style. In real life, many pieces of music usually map to more than one style.

Several methods have been proposed for automatic music style classification~\cite{qin2005music,ZhouZS06,hyper-graph,choi2017convolutional}. Although these methods make some progress, they are limited in two aspects. First, their generalization ability partly suffers from the small quantity of available audio data. Due to the limitation of music copyright, it is difficult to obtain all necessary audio materials to classify music styles. Second, for simplification, most of the previous studies make a strong assumption that a piece of music has only one single style, which does not meet the practical needs. 
% Furthermore, these style categories tend to be too coarse for real-life applications.

%传统的是根据audio information来进行分类的, 由于版权问题，不可能所有的歌曲都有audio information。之前的方法是基于single-label的。
% zgy: 感觉最后一个承接词用Consequently/Therefore 比较 合适。因为是由前面两个reasons 导出的。 如果不是的话，逻辑对于 new reader 也有些突兀

Different from the existing methods, this work focuses on review-driven multi-label music style classification. The motivation of using reviews comes from the fact that, there is a lot of accessible user reviews on relevant websites. First, such reviews provide enough information for effectively identifying the style of music, as shown in Table~\ref{example}. Second, compared with audio materials, reviews can be obtained much more easily. Taking practical needs into account, we do not follow the traditional single-label assumption. Instead, we categorize music items into fine-grained styles and formulate this task as a multi-label classification problem. For this task, we build a new dataset which contains over 7,000 samples. Each sample includes a music title, a set of human annotated styles, and associated reviews. An example is shown in Table~\ref{example}.

% The major challenge of this task lies in the complicated interrelations among music styles, e.g. Classical Music and Piano Music, Electronic Music and Heavy Metal Music, etc. 这一段是不是该再讲一下label之间的关系对结果是什么影响的Although the music styles in each pair are different, there are quite a lot of similarities between them.前面这句话是什么意思啊 This challenge causes that many multi-label classification methods perform badly when applied to this task. Therefore, to better exploit style correlations, we propose a novel deep learning approach with two parts: a label-graph based neural network for modeling, and a new loss function with an interrelation-based continuous label representation for training. 这里的modeling和training是什么意思

%However, if well learned, such relation information is useful for improving the accuracy of predictions. 
% for R\&B, Jazz, Soul Music combining elements of R\&B and Jazz, these music styles not only can be viewed as mutually exclusive labels but also can co-occur in different samples. Therefore, there are different possible label sets, including
% These three style labels can occur independently or co-occur. There are different combinations, such as [R\&B], [Jazz], [Soul Music], [Soul Music, R\&B],  [Soul Music, Jazz], [Soul Music, R\&B, Jazz]. 

%soul包含R&B和Jazz。但是因为soul和R&B又可以单独使用，又可以组合使用，使得模型很难学到soul包含R&B和Jazz这个关系，那么就很容易把soul音乐类型识别成 R&B和Jazz。

%%%%%%%%%%%%%%%%%%%%%%%%%%%%%%%%%%%%%%%%%%%%%%%%
\begin{table*}[t]
\centering
\footnotesize
    \begin{tabular}{|c|p{0.8\linewidth}|}
    \hline
    Music Title  &  	Mozart: The Great Piano Concertos, Vol.1 \\
    \hline
    Styles & Classical Music, Piano Music\\
    \hline
    \multirow{5}{*}{Reviews}& \textsl{(1) I've been listening to \textbf{classical} music all the time.}\\
    %&\textsl{Mozart, the spirit of the music sent by God.}\\
    &\textsl{(2) Mozart is always good. There is a reason he is ranked in the top 3 of lists of greatest \textbf{classical} composers.}\\
    &\textsl{(3) The sound of \textbf{piano} brings me peace and relaxation. }\\
    &\textsl{(4) This volume of Mozart concertos is superb.}\\
    \hline  
    \end{tabular}
    \caption{An illustration of review-driven multi-label music style classification. For easy interpretation, we select a simple and clear example where styles can be easily inferred from reviews. In practice, the correlation between styles and associated reviews is relatively complicated. }
    \label{example}
\vspace{-1\baselineskip}
\end{table*}
%The type of classical music can be inferred from ``listening to classical music'', ``other classical composers''. The type of piano music can be inferred from ``The sound of piano brings me peace and relaxation''.
%%%%%%%%%%%%%%%%%%%%%%%%%%%%%%%%%%%%%%%%%%%%%%%%%%%%%%
The major challenge of this task lies in the complicated correlations of music styles. For example, Soul Music\footnote[1]{Soul Music is a popular music genre that originated in the United States in the late 1950s and early 1960s. It contains elements of African-American Gospel Music, R\&B and Jazz. } contains elements of R\&B and Jazz. These three labels can be used alone or in combination. Many multi-label classification methods fail to capture this correlation, and may mistake the true label [Soul Music, R\&B, Jazz] for the false label [R\&B, Jazz]. If well learned, such relations are useful knowledge for improving the performance, e.g., increasing the probability of Soul Music if we find that it is heavily linked with two high probability labels: R\&B and Jazz. Therefore, to better exploit style correlations, we propose a novel deep learning approach with two parts: a label-graph based neural network, and a soft training mechanism with correlation based continuous label representation.

First, the label-graph based neural network is responsible for classifying music styles based on reviews and style correlations. A hierarchical attention layer collects style-related information from reviews based on a two-level attention mechanism, and a label graph explicitly models the relations of styles. Two information flows are combined together to output the final label probability distribution.

Second, we propose a soft training mechanism by introducing a new loss function with continuous label representation that reflects style correlations. Without style relation information, the traditional discrete label representation sometimes over-distinguishes correlated styles, which does not encourage the model to learn style correlations and limits the performance. Suppose a sample has a true label set [Soul Music], and currently the output probability for Soul Music is 0.8, and the probability for R\&B is 0.3. It is good enough to make a correct prediction of [Soul Music]. However, the discrete label representation suggests the further modification to the parameters, until the probability of Soul Music becomes 1 and the probability of R\&B becomes 0. Because Soul Music and R\&B are related as mentioned above, over-distinguishing is harmful for the model to learn the relation between Soul Music and R\&B. To avoid this problem, we introduce the continuous label representation as the supervisory signal by taking style correlations into account. Therefore, the model is no longer required to distinguish styles completely because a soft classification boundary is allowed.

Our contributions are the followings:
\begin{itemize}

\item To the best of our knowledge, this work is the first to explore review-driven multi-label music style classification.\footnote[2]{The dataset is in the supplementary material and we will release it if this paper is accepted.}

\item To learn the relations among music styles, we propose a novel deep learning approach with two parts: a label-graph based neural network, and a soft training mechanism with correlation-based continuous label representation. 

\item Experimental results on the proposed dataset show that our approach achieves significant improvements over the baselines in terms of all evaluation metrics. 

\end{itemize}
\section{Related works}
%This paper is related with music style classification and multi-label classification. In this section, we give a detailed introduction about the related studies.

\subsection{Music Style Classification}

Previous works mainly focus on using audio information to identify music styles. Traditional machine learning algorithms are adopted in this task, such as Support Vector Machine (SVM)~\cite{xu2003musical}, Hidden Markov Model (HMM)~\cite{chai2001folk,DBLP:journals/taslp/PikrakisTK06}, and Decision Tree (DT)~\cite{ZhouZS06}. Furthermore, several studies explore different  hand-craft feature templates~\cite{tzanetakis2002musical,qin2005music,review}.
% \newcite{tzanetakis2002musical} proposed three feature sets to represent timbral texture, rhythmic content, and pitch content. \newcite{qin2005music} took audio as data source and mined frequent patterns of different music. \newcite{review} extract a variety of features to perform music genre classification. 
Recently, neural networks have freed researchers from cumbersome feature engineering. For example, \citet{choi2017convolutional} introduced a convolutional recurrent neural network for music classification. \citet{medhat2017automatic} designed a masked conditional neural network for multidimensional music classification.

Motivated by the fact that many pieces of music usually have different styles, several studies aim at multi-label musical style classification. For example, \citet{hyper-graph} proposed to solve multi-label music genre classification with a hyper-graph based SVM. \newcite{oramas2017multi} explored how representation learning
approaches for multi-label audio classification outperformed traditional handcrafted feature based approaches. 

The previous studies have two limitations. First, they are in shortage of available audio data, which limits the generalization ability. Second, their studies are based on a strong assumption that a piece of music should be assigned with only one style. Different from these studies, we focus on using easily obtained reviews in conjunction with multi-label music style classification.

\subsection{Multi-Label Classification}

In contrast to traditional supervised learning, in multi-label learning, each music item is associated with a set of labels. Multi-label learning has gradually attracted attention, and has been widely applied to diverse problems, including image classification~\cite{DBLP:conf/mm/QiHRTMZ07,DBLP:conf/civr/WangZC08}, audio classification~\cite{DBLP:journals/pr/BoutellLSB04,DBLP:conf/sigir/SandenZ11}, web mining~\cite{DBLP:conf/nips/KazawaITM04}, information retrieval~\cite{DBLP:conf/sigir/ZhuJXG05,DBLP:conf/sigir/GopalY10}, etc. Compared to the existing multi-label learning methods~\cite{DBLP:journals/corr/abs-1805-04033, DBLP:journals/corr/abs-1801-09030,DBLP:journals/corr/abs-1808-05437,DBLP:conf/coling/YangSLMWW18}, our method has novelties: a label graph that explicitly models the relations of styles; a soft training mechanism that introduces correlation-based continuous label representation. To our knowledge, most of the existing studies of learning label representation only focus on single-label classification~\cite{DBLP:journals/corr/HintonVD15,label}, and there is few research on multi-label learning. %The key idea of these studies is to explore the label correlations. Following these studies, we propose a novel method for review-driven multi-label music style classification that contains much more complicated label relations than traditional multi-label classification task. The proposed method consists of two parts: a label-graph based neural network and a soft training mechanism by learning correlation-based label representations. To our knowledge, most of the existing studies of learning label representations only focus on single-label classification~\cite{DBLP:journals/corr/HintonVD15,label}, and there is few research on multi-label learning.
\section {Review-Driven Multi-Label Music Style Classification}

\subsection{Task Definition}

Given several reviews from a piece of music, the task requires the model to predict a set of music styles. Assume that $X = \{\boldsymbol{x}_{1}, \ldots, \boldsymbol{x}_{i}, \ldots, \boldsymbol{x}_{K}\}$ denotes the input $K$ reviews, and $\boldsymbol{x}_{i}={x_{i,1},\ldots, x_{i,J}}$ represents the $i^{th}$ review with $J$ words. The term $Y=\{y_{1}, y_{2}, \ldots, y_{M}\}$ denotes the gold set with $M$ labels, and $M$ varies in different samples. The target of review-driven multi-label music style classification is to learn the mapping from input reviews to style labels.

\subsection{Dataset}
We construct a dataset consisting of 7172 samples. The dataset is collected from a popular Chinese music review website,\footnote[3]{https://music.douban.com} where registered users are allowed to comment on all released music albums. 

The dataset contains 5020, 646, and 1506 samples for training, validation, and testing respectively. We define an album as a data sample in the dataset, the dataset contains over 287K reviews and over 3.6M words.  22 styles are found in the dataset.\footnote[4]{The styles include: Alternative Music, Britpop, Classical Music, Country Music, Dark Wave, Electronic Music, Folk Music, Heavy Metal Music, Hip-Hop, Independent Music, Jazz, J-Pop, New-Age Music, OST, Piano Music, Pop, Post-Punk, Post-Rock, Punk, R\&B, Rock, and Soul Music.} Each sample is labeled with 2 to 5 style types.
Each sample includes the title of an album, a set of human annotated styles, and associated user reviews sorted by time. An example is shown in Table~\ref{example}. 
% For efficiency, we only collect the top 40 reviews for each data sample, and those samples with less than 40 reviews are dropped. 
On average, each sample contains 2.2 styles and 40 reviews, each review has 12.4 words.

\begin{figure*}
\centering
\includegraphics[width=0.95\linewidth]{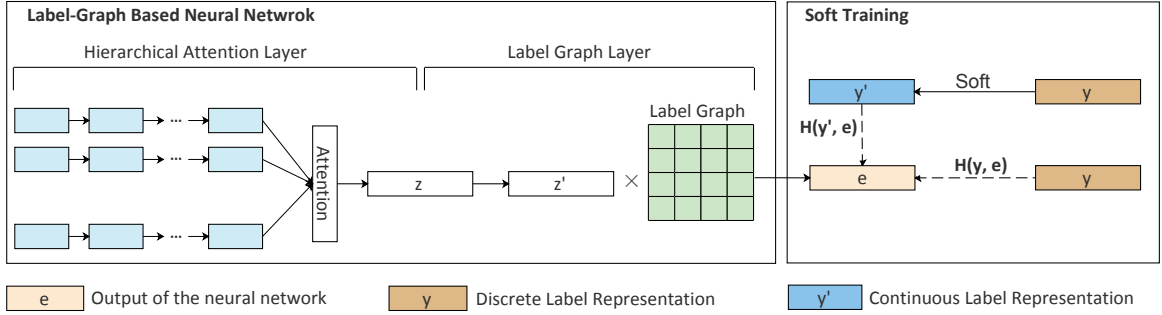}
\caption{An illustration of the proposed approach. Left: The label-graph based neural network. Right: The soft training method. The label graph defines the relations of labels. $\boldsymbol{e}$ is the output label probability distribution. Soft training means that we combine the continuous label representation $\boldsymbol{y}'$ and the discrete label representation $\boldsymbol{y}$ together to train the model. The hierarchical attention layer is responsible for extracting style-related information. The label graph layer and soft training are used for exploiting label correlations. }
\label{fig:hie}
\end{figure*}

\section{Proposed Approach}

In this section, we introduce our proposed approach in detail. An overview is presented in Section~\ref{overview}. The details are explained in Section~\ref{model} and Section~\ref{soft}. 

\subsection{Overview}\label{overview}

The proposed approach contains two parts: a label-graph based neural network and a soft training mechanism with continuous label representation. An illustration of the proposed method is shown in Figure~\ref{fig:hie}.  

The label-graph based neural network outputs a label probability distribution $e$ based on two kinds of information: reviews and label correlations. First, a hierarchical attention layer produces a music representation $\boldsymbol{z}$ by using a two-level attention mechanism to extract style-related information from reviews. Second, we transforms $\boldsymbol{z}$ into a ``raw'' label probability distribution  $\boldsymbol{z}'$  via a sigmoid function. Third, a label graph layer outputs the final label probability distribution $\boldsymbol{e}$ by multiplying the ``raw'' label representation with a label graph that explicitly models the relations of labels. Due to noisy reviews, the model sometimes cannot extract all necessary information needed for a correct prediction. The label correlations can be viewed as supplementary information to refine the label probability distribution. For example, the low probability of a true label will be increased if the label is heavily linked with other high probability labels. With the label correlation information, the model can better handle multi-label music style classification, where there are complicated correlations among music styles. 

Typically, the model is trained with the cross entropy between the discrete label representation $\boldsymbol{y}$ and the predicted label probability distribution  $\boldsymbol{e}$. However, we find it hard for the model to learn style correlations because the discrete label representation does not explicitly contain style relations. For example, for a true label set [Soul Music], the discrete label representation assigns Soul Music with  the value of 1 while its related styles, R\&B and Jazz, get the value of 0. Such discrete distribution does not encourage the  model to learn the relation between Soul Music and its related styles. To better learn label correlations, a continuous label representation $\boldsymbol{y}'$ that involves label relations is desired as training target. Therefore, we propose a soft training method that combines the traditional discrete label representation $\boldsymbol{y}$ (e.g., $[1, 1, 0]$) and the continuous label representation $\boldsymbol{y}'$ (e.g., $[0.80, 0.75, 0.40]$).

We first propose to use the learned label graph $\mathcal{G}$ to transform the discrete representation $\boldsymbol{y}$ into a continuous form. The motivation comes from that in a well-trained label graph, the values should reflect label relations to a certain extent. Two highly related labels should get a high relation value, and two independent labels should get a low relation value. However, in practice, we find that for each label, the relation value with itself is too large and the relation value with other labels is too small, e.g., [0.95, 0.017, 0.003]. It causes the generated label representation  lacking sufficient label correlation information. Therefore, to enlarge the label correlation information in the generated label representation, we propose a smoothing method that punishes the high relation values and rewards the low relation values in $\mathcal{G}$. The method applies a softmax function with a temperature $\tau$ on $\mathcal{G}$ to get a softer label graph $\mathcal{G}'$, and uses $\mathcal{G}'$ to transform $\boldsymbol{y}$ into a softer label representation.

For ease of understanding, we introduce our approach from the following two aspects: one for extracting music representation from reviews, the other for exploiting label correlations.

\subsection{Hierarchical Attention Layer for Extracting Music Representation}\label{model}

This layer takes a set of reviews $X$ from the same sample as input, and outputs a music representation $\boldsymbol{z}$. Considering that the dataset is built upon a hierarchical structure where each sample has multiple reviews and each review contains multiple words, we propose a hierarchical network to collect style-related information from reviews.

We first build review representations via a Bi-directional Long-short Term Memory Network (Bi-LSTM) and then aggregate these review representations into the music representation. The aggregation process also adopts a Bi-LSTM structure that takes the sequence of review representations as input. Second, it is observed that different words and reviews are differently informative. Motivated by this fact, we introduce a two level of attention mechanism~\cite{DBLP:journals/corr/BahdanauCB14}: one at the word level and the other at the review level. It lets the model to pay more or less attention to individual words and sentences when constructing the music representation $\boldsymbol{z}$.  

\subsection{Label Correlation Mechanism}\label{soft}

\subsubsection{Label Graph Layer}
To explicitly take advantage of the label correlations when classifying music styles, we add a label graph layer to the network. This layer takes a music representation $\boldsymbol{z}$ as input and generates a label probability distribution $\boldsymbol{e}$.

First, given an input $\boldsymbol{z}$, we use a sigmoid function to produce a ``raw'' label probability distribution $\boldsymbol{z}'$ as
\begin{equation}
\boldsymbol{z}' = sigmoid(f(\boldsymbol{z})) = \frac{1}{1+e ^{-f(\boldsymbol{z})}} 
\end{equation}
where $f()$ is a feed-forward network. 

%and the edges $\mathcal{E} \in R_{nl \times nl}$ are the similarities between two labels

Formally, we denote $\mathcal{G} \in R_{m \times m}$ as the label graph, where $m$ is the number of labels in the dataset, $\mathcal{G}$ is initialized by an identity matrix. An element $\mathcal{G}[l_{i}, l_{j}]$ is a real-value score indicating how likely the label $l_{i}$ and the label $l_{j}$ are related in the training data. The graph $\mathcal{G}$ is a part of parameters and can be learned by back-propagation. 

Then, given the ``raw'' label probability distribution $ \boldsymbol{z}'$ and the label graph $\mathcal{G}$, the output of this layer is:
\begin{equation}
\boldsymbol{e} = \boldsymbol{z}' \cdot \mathcal{G}
\end{equation}

Therefore, the probability of each label is determined not only by the current reviews, but also by its relations with all other labels. The label correlations can be viewed as supplementary information to refine the label probability distribution. % For example, the probability of a label heavily linked with many high-probability labels will be increased.

\subsubsection{Soft Training}

Given a predicted label probability distribution $\boldsymbol{e}$ and a target discrete label representation $\boldsymbol{y}$, the typical loss function is computed as
\begin{equation}
\label{loss_ye}
L(\theta) = \mathcal{H}(\boldsymbol{y}, \boldsymbol{e}) = -\sum_{i=1}^{m}{y_{i}\log e_{i}} 
\end{equation}
where $\theta$ denotes all parameters, and $m$ is the number of the labels. The function $ \mathcal{H}(,)$ denotes the cross entropy between two distributions. 

However, the widely used discrete label representation does not apply to the task of music style classification, because the music styles are not mutually exclusive and highly related to each other. The discrete distribution without label relations makes the model over-distinguish the related labels. Therefore, it is hard for the model to learn the label correlations that are useful knowledge.

Instead, we propose a soft training method by combining a discrete label representation $\boldsymbol{y}$ with a correlated-based continuous label representation $\boldsymbol{y}'$. The probability values of $\boldsymbol{y}'$ should be able to tell which labels are correct, and the probability gap between two similar labels in $\boldsymbol{y}'$ should not be large. 
With the combination between $\boldsymbol{y}'$ and $\boldsymbol{y}$ as training target, the classification model is no longer required to distinguish styles completely and can have a soft classification boundary. 

A straight-forward approach to produce the continuous label representation is to use the label graph matrix $\mathcal{G}$ to transform the discrete representation $\boldsymbol{y}$ into a continuous form:
\begin{equation}
\boldsymbol{y_c} = \boldsymbol{y} \cdot \mathcal{G}
\end{equation}
We expect that the values in a well-learned label graph should reflect the degree of label correlations. 
% Two related labels should have a high value and two uncorrelated labels should have a low value.
However, in practice, we find that for each label, the relation value with itself is too large and the relation value with other labels is too small. It causes the generated label representation $\boldsymbol{y_c}$ lacking sufficient label correlation information. Therefore, to enlarge the label correlation information in $\boldsymbol{y_c}$, we propose a smoothing method that punishes the high relation values and rewards the low relation values in $\mathcal{G}$. We apply a softmax function with a temperature $\tau$ on $\mathcal{G}$ to get a softer $\mathcal{G}'$ as 
\begin{equation}
(\mathcal{G}')_{ij}= \frac{\exp{[(\mathcal{G})_{ij}/\tau]}}{\sum_{i=1}^{N}\exp{[(\mathcal{G})_{ij}/\tau]}} 
\end{equation}
where $N$ is the dimension of each column in $\mathcal{G}$. This transformation keeps the relative ordering of relation values unchanged, but with much smaller range. The higher temperature $\tau$ makes the steep distribution softer. Then, the desired continuous representation $\boldsymbol{y}'$ is defined as
\begin{equation}
 \boldsymbol{y}' = \boldsymbol{y} \cdot \mathcal{G}'
\end{equation}

\noindent Finally, 
% to train the model
we define the loss function as
\begin{equation}
Loss(\theta) =  \mathcal{H}(\boldsymbol{e}, \boldsymbol{y}) + \mathcal{H}(\boldsymbol{e}, \boldsymbol{y}')
\end{equation}
where the loss $\mathcal{H}(\boldsymbol{e}, \boldsymbol{y})$ aims to correctly classify labels, and the loss $\mathcal{H}(\boldsymbol{e}, \boldsymbol{y}')$ aims to avoid the over-distinguishing problem and to better learn label correlations. 

With the new objective, the model understands not only which labels are correct, but also the correlations of labels. With such soft training, the model is no longer required to distinguish the labels completely because a soft classification boundary is allowed. %With such soft training, the model is no longer required to distinguish the labels completely because a soft classification boundary is allowed.

\section{Experiment}

In this section, we evaluate our approach on the proposed dataset. We first introduce the baselines, the training details, and the evaluation metrics. Then, we show the experimental results and provide the detailed analysis.

\subsection{Baselines}
We first implement the following widely-used multi-label classification methods for comparison. Their inputs are the music representations which are produced by  averaging word embeddings and review representations at the word level and review level respectively.
\begin{itemize}

%\item Decision Tree: One of the widely-used machine learning methods.

\item ML-KNN~\cite{mlknn}: It is a multi-label learning approach derived from the traditional K-Nearest Neighbor (KNN) algorithm. %For each unseen instance, ML-KNN first identifies its $k$ nearest neighbors in the training set. Based on statistical information gained from the label sets of these neighboring instances, Maximum-a-Posteriori (MAP) principle is used to determine the label set for the unseen instance.

\item Binary Relevance~\cite{DBLP:reference/dmkdh/TsoumakasKV10}: It decomposes a multi-label learning task into a number of independent binary learning tasks (one per class label). It learns several single binary models without considering the dependences among labels.
 
\item Classifier Chains~\cite{read2011classifier}:  It takes label dependencies into account and keeps the computational efficiency of the binary relevance method.

\item Label Powerset~\cite{DBLP:conf/ecml/TsoumakasV07}: All classes assigned to an example are combined into a new and unique class in this method.

\item MLP: It feed the music representations into a multilayer perceptron, and generate the probability of music styles through a sigmoid layer.
\end{itemize}
Different from the above baselines, the following two directly process word embeddings. Similar to MLP, they produce label probability distribution by a feed-forward network and a sigmoid function.
\begin{itemize}
\item CNN: It consists of two layers of CNN which has multiple convolution kernels, then feed the word embeddings to get the music representations.

\item LSTM:  It consists of two layers of LSTM, which processes words and sentences separately to get the music representations.

\end{itemize}

%Furthermore, we also implement the  traditional models (e.g., SVM, Decision Tree), and the neural   
%networks (e.g., MLP, Bi-LSTM) for comparison.

\subsection{Training Details}
The features we use for the baselines and the proposed method are the pre-trained word embeddings of reviews. For evaluation, we introduce a hyper-parameter $p$, and a label  will be considered a music style of the song if its probability is greater than $p$.
We tune hyper-parameters based on the performance on the validation set. We set the temperature $\tau$ in soft training to 3, $p$ to 0.2, hidden size to 128, embedding size to 128, vocabulary size to 135K, learning rate to 0.001, and batch size to 128. The optimizer Adam~\cite{DBLP:journals/corr/KingmaB14}  and the maximum training epoch is set to 100. %Following the work of \newcite{label}, we use the decoupled output mechanism.
We choose parameters with the best performance on the validation set and then use the selected parameters to predict results on the test set.

%Following the work of \newcite{label}, we also apply the decoupled output layers in this work for a better performance. For MLP and Bi-LSTM, the hyper-parameter settings are the same with the proposed method.

% For Label Powerset and Binary Relevance, we use the Support Vector Machine (SVM) as a single binary model. The SVM is trained with a liner kernel and the complexity constant $C$ is set to 120. For MLKNN the number of neighbors is set to 10. 

\subsection{Evaluation Metrics}

Multi-label classification requires different evaluation metrics from traditional single-label classification. In this paper, we use the following widely-used evaluation metrics. 

%Suppose $T$ denotes the true set of labels for a given sample, and $P$ denotes the predicted set of labels. Precision is $ {|T\cap P|}/{|P|}$, recall is ${|T\cap P|}/{|T|}$, and F-score is their harmonic mean. 
%Suppose $T$ denotes the true set of labels for a given sample, and $P$ denotes the predicted set of labels. Precision is $ {|T\cap P|}/{|P|}$, recall is ${|T\cap P|}/{|T|}$, and F-score is their harmonic mean.  
\begin{itemize}

\item F1-score: We calculate the micro F1 and macro F1, respectively.  Macro F1 computes the metric independently for each label and then takes the average, whereas micro F1  aggregates the contributions of all labels to compute the average metric.  %In a multi-class classification setup, micro F-score is preferable if label is imbalanced (i.e you may have many more examples of one class than of other classes)

%Micro F-score is the harmonic mean of micro-averaged Recall and Precision, which are calculated from individual true positives, false positives, and false negatives for different sets. Macro F-score is the harmonic mean of macro-averaged Recall and Precision, which just take the average of the precision and recall of the system.

\item One-Error: One-error evaluates the fraction of examples whose top-ranked label is not in the gold label set.

\item Hamming Loss: Hamming loss counts the fraction of the wrong labels to the total number of labels.
\end{itemize}

\subsection{Experimental Results}
\begin{table}[h]
	\centering
    \footnotesize
     \setlength{\tabcolsep}{2pt}
	\begin{tabular}{l |c| c| c| c }
		\hline
		\multicolumn{1}{l|}{\textbf{Models}} &
        \multicolumn{1}{c|}{\textbf{OE(-)}} & 
		\multicolumn{1}{c|}{\textbf{HL (-)}}&
        \multicolumn{1}{c|}{\textbf{Macro F1(+)}} &
         \multicolumn{1}{c}{\textbf{Micro F1(+)}}    \\ \hline
        ML-KNN  &77.3 &0.094 &23.6 &38.1 \\
		Binary Relevance  &74.4 &0.083&24.7 &41.8   \\ 
		Classifier Chains  &67.5 &0.107&29.9&44.3     \\
		Label Powerset  &56.2  &0.096  &37.7 &50.3 \\ 
        MLP  &71.5 &0.081  &29.8&45.8 \\
        CNN &37.9 &0.099 &32.5 &49.3	\\
        LSTM &30.5 &0.089 &33.0  &53.9 \\ \hline %	p=0.3
%         GRU & & & & 	\\ \hline
        \textbf{HAN (Proposal)} &25.9 &0.079  &52.1 & 61.0	\\
\textbf{+LCM (Proposal)} & \textbf{22.6} & \textbf{0.074}& \textbf{54.4}& \textbf{64.5}\\ \hline %p=0.2
	\end{tabular}
	\caption{The comparisons between our approach and the baselines on the test set. The OE and HL denotes one-error and hamming loss respectively, the implemented approach HAN and LCM denotes the hierarchical attention network and the label correlation mechanism respectively. ``+'' represents that higher scores are better and ``-'' represents that lower scores are better.  It can be seen that the proposed approach significantly outperforms the baselines. }
    \label{table:state-of-the-art}
\end{table}

We evaluate our approach and the baselines on the test set. The results are summarized in Table~\ref{table:state-of-the-art}. It is obvious that the proposed approach significantly outperforms the baselines, with micro F1 of 64.5, macro F1 of 54.4, and one-error of 22.6, improving the metrics by 10.6, 21.4, and 7.9 respectively. The improvement is attributed to two parts, a hierarchical attention network and a label correlation mechanism. Only using the hierarchical attention network outperforms the baselines, which shows the effectiveness of hierarchically paying attention to different words and sentences. The greater F1-score is achieved by adding the proposed label correlation mechanism, which shows the contribution of exploiting label correlations. Especially, the micro F1 is improved from 61.0 to 64.5, and the macro F1 is improved from 52.1 to 54.4. 
%, which demonstrates that the proposed label-correlations mechanism helps a lot by taking advantage of label correlations. 

The results of baselines also reveal the usefulness of label correlations for improving the performance. ML-KNN and Binary Relevance, which over-simplify multi-label classification and neglect the label correlations, achieve the worst results. In contrast, Classifier Chains and Label Powerset, which take label correlations into account, get much better results. Though without explicitly taking advantage of label correlations, the neural baselines, MLP, CNN, and LSTM, still achieve better results, due to the strong learning ability of neural networks.

\subsection{Incremental Analysis}
\begin{table}[h]
	\centering
    \footnotesize
	\begin{tabular}{l| c| c| c| c }
		\hline
		\multicolumn{1}{l|}{\textbf{Models}} &
         \multicolumn{1}{c|}{\textbf{OE(-)}} & 
         \multicolumn{1}{c|}{\textbf{HL(-)}}  &
         \multicolumn{1}{c|}{\textbf{Macro F1(+)}} &
         \multicolumn{1}{c}{\textbf{Micro F1(+)}} 
		   \\ \hline
           
           % LSTM &30.5 &0.089 &33.0  &53.9 \\ \hline %	p=0.3
        
        HAN &25.9 &0.079  &52.1 & 61.0	\\	%p=0.3
        +\textbf{LG} &23.4 &0.077 &54.2  &62.8   \\  %p=0.3
%         + Soft Training ($\boldsymbol{e'}$)&25.1&\textbf{0.070}&57.9 &42.9\\ %p=0.4 
        %+ Soft Training ($\boldsymbol{e}$) &22.4&0.083&53.5&62.6  \\ %p=0.4 
%        + Soft Training ($\boldsymbol{y_c}$) &&&&\\ %p =0.35 1e-3
%         + Soft Training ($\boldsymbol{y}'$)&22.4& \textbf{0.071}&\textbf{65.7}&\textbf{56.6}\\ \hline %p=0.2     
        + \textbf{ST} & \textbf{22.6} & \textbf{0.074}& \textbf{54.4}& \textbf{64.5}\\ \hline %p=0.2
	\end{tabular}
	\caption{Performance of key components in the proposed approach. LG and ST denote the label graph layer and the soft training.}
    \label{Tab:module}
\end{table}
In this section, we conduct a series of experiments to evaluate the contributions of our key components. The results are shown in Table~\ref{Tab:module}.

%From the results, we can see that the method with the label graph layer achieves large improvements by explicitly modeling the correlation between two labels, with a 7.2 micro F-score improvement and a 22.7 macro F-score improvement. It shows that the label graph is an effective mechanism in taking advantage of label correlations. Furthermore, the proposed soft training method also brings the better improvement by introducing an interrelation-based continuous label representation that helps the traditional training method in teaching the label graph to better learn the label correlations. Specially, the micro F-score is improved from 62.8 to 65.3 and the macro F-score is improved from 54.2 to 57.6. 

%However, since the traditional training method with an one-hot label representation only reflects which labels are correct and neglects the relations between the correct label and the incorrect labels, causing that the label graph only learns the limited label correlations. Therefore, to solve this problem, we propose a soft training method. 

%with an one-hot label representation usually encourages the model to over-distinguish the labels which are related with each other, which limits the model to learn label correlations
The method with the label graph does not achieve the expected improvements. It indicates that though with explicitly modeling the label correlations, the label graph does not play the expected role. It verifies our assumption that the traditional training method with discrete label representation makes the model over-distinguish the related labels, and thus does not learn label correlations well. To solve this problem, we propose a soft training method with a continuous label representation $\boldsymbol{y}'$ that takes label correlations into account. It can be clearly seen that with the help of soft training, the proposed method achieves the best performance. Especially, the micro F-score is improved from 62.8 to 64.5, and the one-error is reduced from 23.4 to 22.6.  With the new loss function, the model not only knows how to distinguish the right labels from the wrong ones, but also can learn the label correlations that are useful knowledge, especially when the input data contains too much 
style unrelated words for the model to extract all necessary information.   

%We also explore different kinds of continuous label representations. The method that uses $\boldsymbol{y_c}$ as the soft label representation does not beat the method that uses $\boldsymbol{y}'$. It is mainly because the value of $\boldsymbol{y_c}$ contains little label correlation information. In contrast, the label relation information is enlarged in $\boldsymbol{y}'$ and thus guides the model to better learn label correlations. 
\begin{table}[h]
	\centering
    \footnotesize
    \setlength{\tabcolsep}{2pt}
	\begin{tabular}{ p{0.3\linewidth}|p{0.33\linewidth}|p{0.3\linewidth} }
		\hline
		\textbf{Ground Truth} &
        \textbf{Without LCM}  &
		\textbf{With LCM} \\ \hline

      Britpop\footnotemark[5], Rock 
      &\textcolor{blue}{Britpop}& \textcolor{blue}{Britpop}, \textcolor{orange}{Rock}
      \\ 
      \hline 
%       Country Music, Pop & \textcolor{blue}{Pop} & \textcolor{blue}{Pop}, \textcolor{purple}{Country Music}	\\ \hline
%       Folk, Independent Music & \textcolor{blue}{Folk} &\textcolor{blue}{Folk}, \textcolor{purple}{Independent Music}	\\ \hline % name: Elliott Smith  0.67,0.31  Han:0.9, a dozen of .1,0.2
      Hip-Hop\footnotemark[6], Pop, R\&B\footnotemark[7] &Electronic Music, \textcolor{blue}{Pop}&
      \textcolor{blue}{Pop}, \textcolor{orange}{R\&B}\\ \hline % name: Rihanna Pop0.52,r&b0.39 Han:0.96,0.95
      Pop, R\&B & \textcolor{blue}{Pop}, Rock, Britpop & \textcolor{blue}{Pop}, \textcolor{orange}{R\&B} \\ \hline	%name Tension pop0.5,r&b0.2, Han:0.9,0.6,0.3
      Country Music, Folk, Pop & \textcolor{blue}{Country Music}, \textcolor{blue}{Pop} &
      \textcolor{blue}{Country Music}, \textcolor{blue}{Pop}, \textcolor{orange}{Folk}\\ \hline % name Ballads
      Classical Music, New-Age Music\footnotemark[8], Piano Music & \textcolor{blue}{Piano Music}, \textcolor{blue}{Classical Music} & \textcolor{blue}{Piano Music}, \textcolor{orange}{New-Age Music}, \textcolor{blue}{Classical Music} \\ \hline 
       
	\end{tabular}
	\caption{Examples generated by the methods with and without the label correlation mechanism. The labels correctly predicted by two methods are shown in blue. The labels correctly predicted by the method with the label correlation mechanism are shown in orange. We can see that the method with the label correlation mechanism classifies music styles more precisely.}
    \label{Tab:casestudy}
\end{table}
\footnotetext[5]{Britpop is a style of British Rock.}
\footnotetext[6]{Hip-Hop is a mainstream Pop style.}
\footnotetext[7]{Rhythm and Blues, often abbreviated as R\&B, is a genre of popular music.}
% \footnotetext[8]{Independent music is music produced independently from commercial record labels or their subsidiaries, a process that may include an autonomous, do-it-yourself approach to recording and publishing.}
\footnotetext[8]{New-Age Music is a genre of music intended to create artistic inspiration, relaxation, and optimism. It is used by listeners for yoga, massage, and meditation.}
For clearer understanding, we compare several examples generated with and without the label correlation mechanism in Table~\ref{Tab:casestudy}. By comparing gold labels and predicted labels generated by different methods, we find that the proposed label correlation mechanism identifies the related styles more precisely. This is mainly attributed to the learned label correlations. For example, the correct prediction in the first example shows that, the label correlation mechanism captures the close relation between ``Britpop'' and ``Rock'', which helps the model to generate a more appropriate prediction. 
%Furthermore, to analyze why the proposed method works, we visualize the label graph matrix, as shown in Figure~\ref{}. It can be clearly seen that even with any direct signal, the propose methods also make it successfully learn the precious label correlations. 

\subsection{Visualization Analysis}
\label{visualization}
\begin{figure}[h]
\centering
\includegraphics[width=0.8\linewidth]{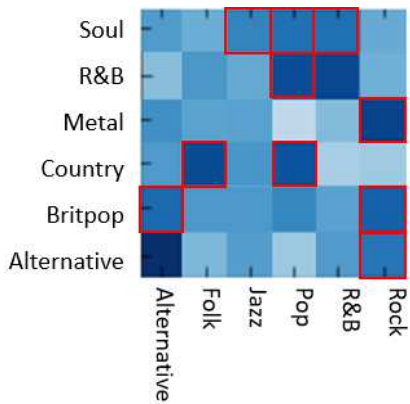}
\caption{The heatmap generated by the learned label graph. The deeper color represents the closer relation. For space, we abbreviate some music style names. We can see that some obvious relations are well captured by the model, e.g., ``Heavy Metal Music (Metal)'' and ``Rock'', ``Country Music (Country)'' and ``Folk''. }
\label{fig:LG}
\end{figure}

Since we do not have enough space to show the whole heatmap of all 22 labels, we randomly select part of the heatmap to visualize the learned label graph. 
Figure \ref{fig:LG} shows that some obvious music style relations are well captured. For ``Country Music'', the most related label is ``Folk Music''. In reality, these two music styles are highly similar and the boundary between them is not well-defined.  For three kinds of rock music, ``Heavy Metal Music'', ``Britpop Music'', and ``Alternative Music'', the label graph correctly captures that the most related label for them is ``Rock''. For a more complicated relation where ``Soul Music'' is highly linked with two different labels, ``R\&B'' and ``Jazz'', the label graph also correctly capture such relation. These examples demonstrate that the proposed approach performs well in capturing  relations among music styles. 

% \footnotetext[8]{Soul music is a popular music genre. It combines elements of African-American gospel music, R\&B and jazz.}

%We also notice that these relations occur in the training data with high frequency. It demonstrates that our approach performs well in capturing label relations.
% Some music styles are associated with several music styles, e.g. Independent Music, Pop,    compared to the niche music styles, these labels are more often marked by users. 
% We can also see that the diagonal elements in the label graph is less than zero, which means that after the soft training, the model becomes less confident to its original prediction but consider more about the effect of other labels and their relationship on the prediction result.

\subsection{Error Analysis}
\label{error}
Although the proposed method has achieved significant improvements, we also notice that there are some failure cases. In this section, we give the detailed error analysis.

First, the proposed method performs worse on the styles with low frequency in the training set. Table ~\ref{Tab:label unbalance} compares the performance on the top 5 music styles of highest and lowest frequencies. As we can see, the top 5 fewest music styles get much worse results than top 5 most music styles. This is because the label distribution is highly imbalanced where unpopular music styles have too little training data. For future work, we plan to explore various methods to handle this problem. For example, re-sample original data to provide balanced labels.
%or use ensemble techniques for reducing over-fitting in labels with high 是低的么？ frequency. 

Second, we find that some music items are wrongly classified into the styles that are similar with the gold styles. For example, a sample with a gold set [Country Music] is wrongly classified into [Folk] by the model. The reason is that some music styles share many common elements and only subtly differ from each other. It poses a great challenge for the model to distinguish them. For future work, we would like to research how to effectively address this problem.
%Second, we find that some music are wrongly classified into the similar music styles. It is d %This is probably the side-effect of the proposed method. The model sometimes over depends on the style correlation information and neglects some distinguishing information. How to balance these two kinds of information also needs to be explored in future work.
%\note{mark1}

% \begin{table}[h]
% 	\centering
%     \footnotesize
% 	\begin{tabular}{l c c | l c c}
% 		\hline
% 		\multicolumn{1}{l}{\textbf{Most Styles}} &
% 		\multicolumn{1}{c}{\textbf{\% of Samples}} &
%         \multicolumn{1}{c}{\textbf{F1}} & \multicolumn{1}{|l}		{\textbf{Least Styles}} &
% \multicolumn{1}{c}{\textbf{\% of Samples}} &
%         \multicolumn{1}{c}{\textbf{F1}} \\ \hline
% %         & & &Jazz&4.3\% &37.5\%	\\ 
% %         & & &Jazz&3.9\% &55.6\%	\\ 
%         Rock &30.4  &75.8 	&Jazz &4.3 &37.5 \\
%         Independent Music &30.0  &64.8 &Heavy Metal Music&3.9 &55.6 \\
%         Pop &26.2 &67.1 &	Hip-Hop &3.1 &7.5		\\
%         Folk Music &21.9 &73.7 &	Post-punk &2.5  &17.1\\
%         Electronic Music &13.9 &61.8  & Dark Wave &1.3 &17.4 \\
% % 		& & &Jazz&4.3\% &37.5\%	\\ 
% %         & & &Jazz&3.9\% &55.6\%	\\ 
%         \hline	
        
% 	\end{tabular}
% 	\caption{ The performance of the proposed method on most and fewest styles.}
%     \label{Tab:label unbalance}
% \end{table}
\begin{table}[h]
	\centering
    \footnotesize
	\begin{tabular}{l c c }
		\hline
		\multicolumn{1}{l}{\textbf{Most Styles}} &
		\multicolumn{1}{c}{\textbf{\% of Samples}} &
        \multicolumn{1}{c}{\textbf{F1}}  \\ \hline
%         & & &Jazz&4.3\% &37.5\%	\\ 
%         & & &Jazz&3.9\% &55.6\%	\\ 
        Rock &30.4  &75.8 	 \\
        Independent Music &30.0  &64.8  \\
        Pop &26.2 &67.1 			\\
        Folk Music &21.9 &73.7 	\\
        Electronic Music &13.9 &61.8    \\ \hline
        \textbf{Least styles} & \textbf{\% of Samples} & \textbf{F1}\\ \hline
        Jazz &4.3 &37.5\\
        Heavy Metal Music&3.9 &55.6\\
        Hip-Hop &3.1 &7.5 \\
        Post-punk &2.5  &17.1\\
        Dark Wave &1.3 &17.4\\ \hline

	\end{tabular}
	\caption{ The performance of the proposed method on most and fewest styles.}
    \label{Tab:label unbalance}
\end{table}
\section{Conclusions}
%In this paper, we propose a new task of mining the style of the record from the record reviews. Based on the analysis of the characteristics of this task and comparison between baselines, we propose a new label soften technique that achieve outstanding results in this task.

In this paper, we focus on classifying multi-label music styles with user reviews. To meet the challenge of complicated style relations, we propose a label-graph based neural network and a soft training mechanism. Experiment results show that our proposed approach significantly outperforms the baselines. Especially, the micro F1 is improved from 53.9 to 64.5, and the one-error is reduced from 30.5 to 22.6. Furthermore, the visualization of label graph also shows that our method performs well in capturing label correlations.
\\ 
%For future work, we plan to explore the label imbalance problem. %Second, we plan to explore a more fine-grained version of music style classification  %, the with micro F-score of 64.5, macro F-score of 54.4, and one-error score of 22.6, improving by 10.6, 21.4, and 7.9 respectively.

%There are several research directions for future work. First, we would like to deal with the label imbalance problem. Second, we also plan to explore how to balance the label correlation information and distinguishing information when classifying music styles. 
%Third, the visualization results in Section~\ref{visualization} show that only obvious correlations between styles are captured. We would also like to research on how to capture more fine-grained label correlations.

% In this paper, we focus on classifying multi-label music styles with review texts. To meet the challenge of label correlation, we propose a label-graph neural network and a soft training mechanism. Experiment results have showed the effectiveness of the proposed approach.

\end{CJK}
\normalem
\bibliography{emnlp2018}
\bibliographystyle{acl_natbib_nourl}

\end{document}